\documentclass[letterpaper,table]{article} 
\usepackage[draft]{aaai25}  
\usepackage{times}  
\usepackage{helvet}  
\usepackage{courier}  
\usepackage[hyphens]{url}  
\usepackage{graphicx} 
\usepackage{natbib}  
\usepackage{caption} 
\usepackage{newfloat}
\usepackage{float}  
\usepackage{listings}
\DeclareCaptionStyle{ruled}{labelfont=normalfont,labelsep=colon,strut=off} 
\floatstyle{ruled}
\newfloat{listing}{tb}{lst}{}
\floatname{listing}{Listing}
\usepackage{fnpct}
\usepackage{siunitx}

\title{Weighted Embeddings for Low-Dimensional Graph Representation\thanks{This work was supported by funding from the pilot program Core--Informatics of the Helmholtz Association (HGF).}}
\author {
    Thomas Bläsius\textsuperscript{\rm 1},
    Jean-Pierre von der Heydt\textsuperscript{\rm 1},
    Maximilian Katzmann\textsuperscript{\rm 1},
    Nikolai Maas\textsuperscript{\rm 1}
}
\affiliations {
    \textsuperscript{\rm 1}Karlsruhe Institute of Technology, Karlsruhe, Germany.\\
    thomas.blaesius@kit.edu, heydt@kit.edu, nikolai.maas@kit.edu, maximilian.katzmann@kit.edu
}

\usepackage{amsmath,amssymb}
\usepackage{mathtools}
\usepackage[textsize=tiny]{todonotes}
\usepackage[
    linesnumbered,
    lined, 
    ruled,
    vlined
]{algorithm2e}
\usepackage{booktabs}
\usepackage{csquotes}
\usepackage{cleveref}  
\usepackage{pgfplots}
\usepackage{microtype} 
\usepgfplotslibrary[groupplots]
\usepackage{layouts}
\usepackage{multirow}

\newcommand{\href}[1]{#1}  

\newcommand{\Oh}{\ensuremath{\mathcal{O}}}

\DeclarePairedDelimiter\abs{\lvert}{\rvert}%
\DeclarePairedDelimiter\norm{\lVert}{\rVert}%

\DeclareMathOperator{\dist}{dist}
\DeclareMathOperator{\Prec}{Prec}
\DeclareMathOperator{\Rec}{Rec}

\newcommand{\Loss}{\ensuremath{\mathcal{L}}}
\newcommand{\Lrep}{\ensuremath{\Loss_{\mathrm{rep}}}}
\newcommand{\Lattr}{\ensuremath{\Loss_{\mathrm{attr}}}}
\newcommand{\set}[1]{\left\lbrace #1\right\rbrace}
\newcommand{\frep}{\ensuremath{f_{\mathrm{rep}}}}
\newcommand{\fattr}{\ensuremath{f_{\mathrm{attr}}}}

\newcommand{\embalg}[1]{\textsc{#1}}
\newcommand{\instance}[1]{\texttt{#1}}
\newcommand{\embedder}{\textsc{WEmbed}\xspace}
\newcommand{\verseemb}{\embalg{Verse}\xspace}
\newcommand{\ftovemb}{\embalg{Force2Vec}\xspace}
\newcommand{\deepwalkemb}{\embalg{DeepWalk}\xspace}
\newcommand{\poincareemb}{\embalg{PoincaréEmbed}\xspace}

\definecolor{darkgreen}{rgb}{0.0, 0.5, 0.0}

\newif\ifenablecomments

\enablecommentstrue

\ifenablecomments
\newcommand{\jp}[1]{\textcolor{orange}{#1}}
\newcommand{\nikolai}[1]{\textcolor{darkgreen}{#1}}
\newcommand{\thomas}[1]{\textcolor{purple}{#1}}
\newcommand{\mycomment}[1]{\textcolor{teal}{#1}}
\else
\newcommand{\jp}[1]{}
\newcommand{\nikolai}[1]{}
\newcommand{\thomas}[1]{}
\newcommand{\mycomment}[1]{}
\fi

\begin{document}

\maketitle


\begin{abstract}
  Learning low-dimensional numerical representations from symbolic data, e.g., embedding the nodes of a graph into a geometric space, is an important concept in machine learning.
  While embedding into Euclidean space is common, recent observations indicate that hyperbolic geometry is better suited to represent hierarchical information and heterogeneous data (e.g., graphs with a scale-free degree distribution).
  Despite their potential for more accurate representations, hyperbolic embeddings also have downsides like being more difficult to compute and harder to use in downstream tasks.
  
  We propose embedding into a \emph{weighted space}, which is closely related to hyperbolic geometry but mathematically simpler.
  We provide the embedding algorithm \embedder and demonstrate, based on generated as well as over 2000 real-world graphs, that our weighted embeddings heavily outperform state-of-the-art Euclidean embeddings for heterogeneous graphs while using fewer dimensions.
  The running time of \embedder and embedding quality for the remaining instances is on par with state-of-the-art Euclidean embedders.
\end{abstract}


\section{Introduction}
\label{sec:setting}

Embeddings are a vital concept in machine learning.
They map complex objects like high-dimensional vectors (e.g., images) or symbolic entities (e.g., words of a language or nodes of a graph) into a low-dimensional vector space.
Downstream tasks utilize these low-dimensional representations by perceiving objects as similar if their vector representations are similar.
A natural way to measure the similarity of two vectors $x, y \in \mathbb R^d$ is to interpret them as points in Euclidean space and use their distance $\norm{x - y}$.
A common alternative is the dot product $x \cdot y$, which is closely related to spherical geometry.
However, inspired by findings of the network-science community \cite{Hyper_Geomet_Compl_Networ-Kriouk10}, it has been observed that some types of graphs are better represented by embedding their nodes into hyperbolic geometry \cite{DBLP:conf/nips/NickelK17, Neural_Embed_Graph_Hyper_Space-Chamb17}.

\begin{figure}[t]
  \centering
  \includegraphics[width=\linewidth]{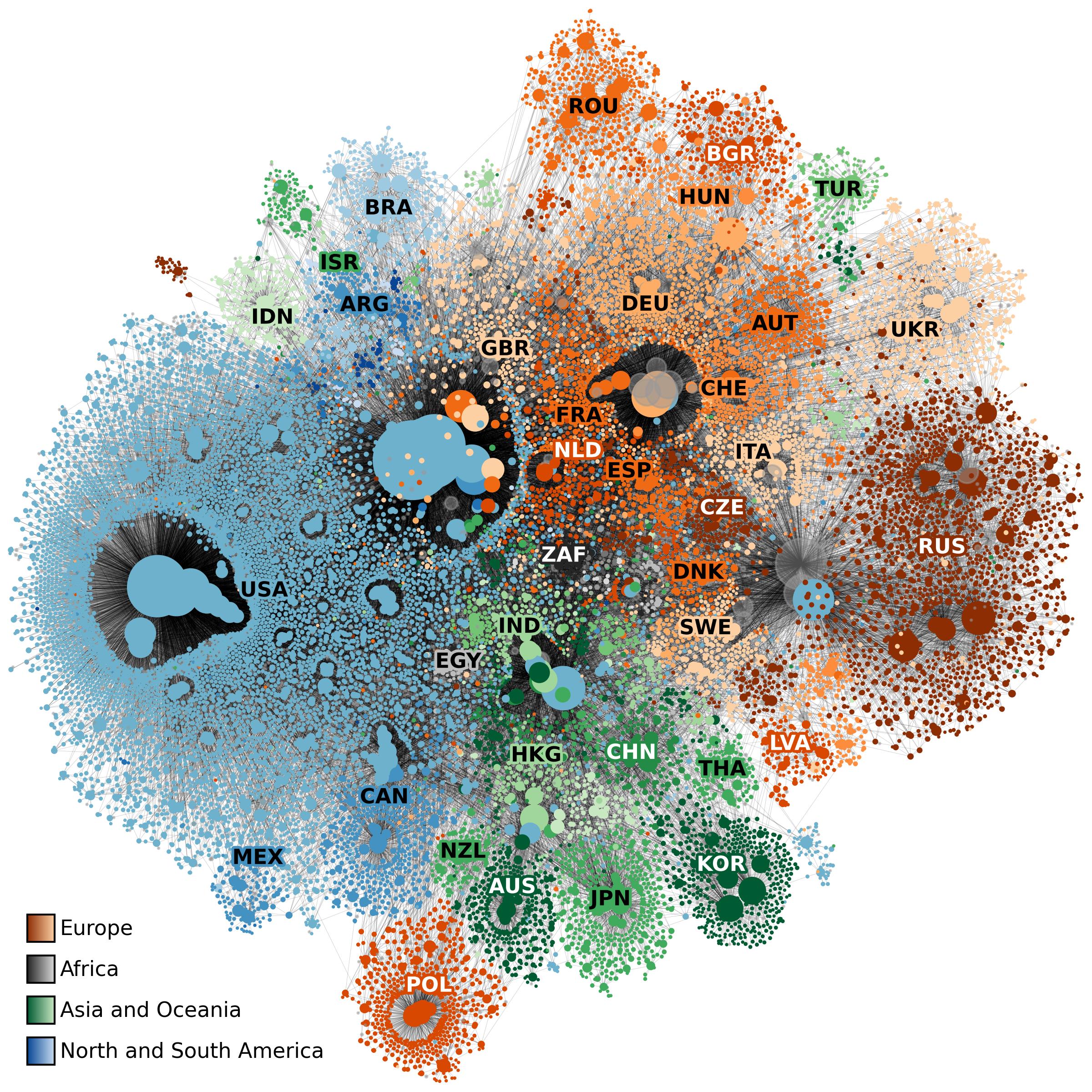}
  \caption{A 2-dimensional weighted embedding of the internet graph used by \citet{Sustain_Inter-Bogun10}.  The node size indicates the weight.}
  \label{fig:internet-2d-emb}
\end{figure}

\paragraph{Benefits of Hyperbolic Geometry}

Hyperbolic space expands exponentially, i.e., the area of a disk of radius $r$ is in $\Theta(e^r)$ for growing $r$.
This makes it possible to embed trees or similar hierarchical structures with low distortion even in the $2$-dimensional hyperbolic plane \cite{Low_Distor_Delaun_Embed_Trees_Hyper_Plane-Sarkar11}.
In comparison, when embedding, e.g., a complete binary tree in the Euclidean plane, one quickly runs out of space as the number of nodes grows exponentially with the layer while the area in the plane only grows quadratically.
Moreover, using hyperbolic distances as similarity measure makes it possible for a node to be similar to many other nodes that are themselves pairwise dissimilar.
This enables low-dimensional representations of heterogeneous graphs\footnote{The term \emph{heterogneous graph} is sometimes also used for graphs with heterogeneous labels.
  In this paper, graphs are always unlabeled and heterogeneity refers to the degree distribution.}.
This in particular includes the prevalent so-called scale-free networks, where the degree distribution follows a power-law \cite{Hyper_Geomet_Compl_Networ-Kriouk10, Scale_Free_Networ_Well_Done-Voital19}.
We note that it is not impossible to find good Euclidean embeddings for hierarchical structures or graphs with heterogeneous degrees but it at least comes at the cost of requiring a significantly higher-dimensional space.

Given these benefits of hyperbolic space when it comes to representing hierarchical or heterogeneous data, it is not surprising that hyperbolic embeddings have gained popularity in recent years.
Although there have been earlier works on embedding graphs in the $2$-dimensional hyperbolic plane \cite{Sustain_Inter-Bogun10, Networ_Mappin_Replay_Hyper_Growt-Papad15, Networ_Geomet_Infer_Using_Common_Neigh-Papad15, Hyper_Mappin_Compl_Networ_Based_Commun_Infor-Wang16, Effic_Embed_Scale_Graph_ESA2016}, the research in the artificial intelligence community has gained traction with the seminal paper by \citet{DBLP:conf/nips/NickelK17}, who demonstrated that hyperbolic embeddings can outperform their Euclidean counterpart for graph reconstruction and link prediction tasks, particularly for low-dimensional embeddings ($d \in \{10, 20\}$).
This result has been extended to and reproduced for, e.g., other coordinate systems in hyperbolic space \cite{DBLP:conf/icml/NickelK18, Loren_Distan_Learn_Hyper_Repres-Law19}, to directed acyclic graphs \cite{Hyper_Entail_Cones_Learn_Hierar_Embed-Ganea18, Hyper_Disk_Embed_Direc_Acycl_Graph-Suzuk19}, or to representing general metrics \cite{DBLP:conf/icml/SalaSGR18}.

We briefly want to mention that hyperbolic geometry has also been incorporated into neural networks \cite{Hyper_Neural_Networ-Ganea18, Hyper_Neural_Networ-Shimiz21} and graph neural networks \cite{Hyper_Graph_Neural_Networ-Liu19, Hyper_Graph_Convol_Neural_Networ-Chami19}, and has been used for other domains such as natural language processing \cite{Poinc_Glove-Tifrea19}, or in computer vision \cite{Hyper_Image_Embed-Khrul20}.
For an overview, see the excellent survey by \citet{Hyper_Deep_Neural_Networ-Peng22} or the collection by \citet{Hyper_Repres_Deep_Learn-Mengl24}.

\paragraph{Challenges with Hyperbolic Geometry}

There are, however, also downsides to using hyperbolic embeddings.
Firstly, hyperbolic geometry is not very accessible and adjusting down-stream tasks to properly interpret hyperbolic coordinates requires quite some mathematical machinery, e.g., \cite{Hyper_Neural_Networ-Ganea18, Hyper_Neural_Networ-Shimiz21}.
There is an effort to mitigate these difficulties, e.g., via tutorials \cite{Hyper_Neural_Networ-Choud22, Hyper_Graph_Neural_Networ-Zhou23}, or by providing easy-to-use libraries \cite{GraphZoo-Vyas22, HypLL-Speng23}.
However, hyperbolic embeddings remain more difficult to use and reason about, in particular for someone not familiar with hyperbolic geometry.

Another issue is that finding good hyperbolic embeddings is algorithmically challenging for multiple reasons.
To explain this, we note that in hyperbolic embeddings, the distance of a node to the origin encodes its centrality in the sense that central nodes have small hyperbolic distance to many other nodes.
Additionally, hyperbolic space behaves like Euclidean space close to the origin and its hyperbolic nature becomes more pronounced further out.
This makes it crucial to choose the right scale for the embedding, which is in stark contrast to Euclidean space where one can scale an embedding without altering relative distances.

As geodesics between points ``bend'' towards the origin, moving a vertex towards a distant neighbor mainly makes it more central, which leads to loss functions with spikes and suboptimal plateaus, making gradient descend less effective \cite{Force_Direc_Embed_Scale_Free-Blasius21}.
Existing embedding methods mitigate this issue, e.g., by having a burn-in phase in which an embedding is computed close to the origin before shifting vertices outwards \cite{DBLP:conf/nips/NickelK17}, by first embedding only a dense core of the graph close to the origin \cite{Sustain_Inter-Bogun10, Effic_Embed_Scale_Graph_ESA2016}, or by optimizing the distance from the origin separately \cite{Force_Direc_Embed_Scale_Free-Blasius21}.
In their survey, \citet{Hyper_Deep_Neural_Networ-Peng22} also point out that different optimizations for stochastic gradient descend are \textquote{designed to optimize parameters living in Euclidean space}.
Additionally, common techniques to reduce the running time, like negative sampling or geometric data structures for low-dimensional embeddings, are less effective or more complicated in hyperbolic space \cite{Gener_Bound_Graph_Embed_Using_Negat_Sampl-Suzuk21, Quadt_Stein_Spann_Approx_Neares-KisfalWordr24}.
Moreover, there are numerical difficulties depending on the used coordinate system \cite{DBLP:conf/icml/NickelK18, Loren_Distan_Learn_Hyper_Repres-Law19, Numer_Stabil_Hyper_Repres_Learn-Mishn23}.

We believe that these challenges pose a real cost that counteracts the superiority of hyperbolic geometry in terms of its representational power.
In fact, in an ideal world without these challenges, there is no reason to favor Euclidean over hyperbolic geometry as a sufficiently small ball in hyperbolic space behaves (almost) equivalent to Euclidean space.
Nonetheless, the consensus in the literature is that hyperbolic embeddings are not always superior but mostly for low dimensions and hierarchical or heterogeneous data\footnote{Although this seems to be the general consensus, there is, to the best of our knowledge, no extensive study on what type of graphs benefit from hyperbolic geometry. Most evaluations are done on just a handful of graphs.} \cite{Hyper_Deep_Neural_Networ-Peng22}.
Moreover, one additionally has to account for the fact that the comparison is usually done with some ad-hoc Euclidean embedding method.
This yields a fair comparison for evaluating the benefits of hyperbolic geometry in isolation.
However, we believe that we are now at a stage where it is crucial to demonstrate usefulness compared to state-of-the-art Euclidean embedders like \embalg{DeepWalk}~\cite{DBLP:conf/kdd/PerozziAS14}, \verseemb~\cite{DBLP:conf/www/TsitsulinMKM18}, and \embalg{Force2Vec}~\cite{DBLP:conf/icdm/RahmanSA20}.
Besides the embedding quality, we believe that running time and ease of use are also important factors where hyperbolic embedding algorithms currently fall behind.

\paragraph{Our Contribution}

Hyperbolic embeddings have been inspired by insights form the network-science community on hyperbolic random graphs~\cite{Hyper_Geomet_Compl_Networ-Kriouk10}.
More recent results by \citet{DBLP:journals/tcs/BringmannKL19} show that hyperbolic random graphs are in some sense equivalent to random graphs from a weighted geometric space.
Motivated by this, we propose \emph{weighted embeddings}, which are defined as follows.
Each node $v$ is mapped to a \emph{position} $p_v \in \mathbb R^{d}$ and additionally to a \emph{weight} $w_v \in \mathbb R^+$.
As a similarity measure between vertices $u$ and $v$, we use the \emph{weighted distance}
\begin{equation}
  \label{eq:weighted-distance}
  \dist(u, v) = \frac{\norm{p_u - p_v}}{(w_uw_v)^{1/d}},
\end{equation}
where $\norm{p_u - p_v}$ is the Euclidean distance between $p_u$ and~$p_v$.
Although these weighted coordinates are not exactly equivalent to hyperbolic coordinates, \citet{DBLP:journals/tcs/BringmannKL19} provide a translation between the two, showing that these weighted spaces are equally capable of representing heterogeneous and hierarchical graphs.

Thus, using weighted distances lets us reap the benefits of hyperbolic space while eliminating some of the challenges coming with it.
Clearly, weighted distances are mathematically very simple, which makes weighted embeddings easier to reason about or to incorporate into downstream tasks.
It is obvious that using uniform weights yields the Euclidean setting as special case.
Moreover, in contrast to hyperbolic geometry, one can scale a weighted embedding arbitrarily, which eliminates the algorithmic difficulty of having to choose the right scale.
Also, optimizing the positions independent from the weights leads to more well-behaved gradients.

Our main contribution is the \embedder algorithm that we engineered to efficiently compute high-quality weighted embeddings.
It works out of the box without any graph-specific parameter tuning, outperforms \embalg{PoincaréEmbed} \cite{DBLP:conf/nips/NickelK17} by orders of magnitude, and is not much slower than the state-of-the-art Euclidean embedding algorithms \embalg{DeepWalk}~\cite{DBLP:conf/kdd/PerozziAS14}, \verseemb~\cite{DBLP:conf/www/TsitsulinMKM18}, and \embalg{Force2Vec}~\cite{DBLP:conf/icdm/RahmanSA20}.
In regards to embedding quality, \embedder heavily outperforms its Euclidean competitors for low-dimensional embeddings of heterogeneous graphs.
For higher dimensions or less heterogeneous networks, \embedder produces comparable results.
To give one example, for the commonly used \instance{citeseer} network, \embedder computes an 8-dimensional embedding in under \SI{11}{s}, which allows almost perfect reconstruction of its edges with an F1 score of \num{0.9999}.
\verseemb, the strongest Euclidean competitor, achieves the scores \num{0.64}, \num{0.977}, and \num{0.992}, in 8, 32, and 128 dimensions, respectively, with slightly higher running times.

Besides evaluating \embedder on commonly used networks, we run experiments on generated graphs to evaluate its scaling behavior, and to gain insights into how weighted and Euclidean embeddings compare depending on the heterogeneity of the graph.
Moreover, we evaluate \embedder and its Euclidean competitors on a set of over 2000 real-world graphs, showing that \embedder consistently produces high-quality embeddings.
\embedder is outperformed only slightly on few graphs but produces substantially better results on many graphs, particularly those with high heterogeneity.
This demonstrates that \embedder is often the best and generally a good choice for computing low-dimensional representations.

Our core contributions can be summarized as follows.
\begin{itemize}
\item We propose weighted embeddings as a way to get the benefits from hyperbolic embeddings without the mathematical and algorithmic challenges coming with it.
\item We engineer the efficient embedding algorithm \embedder and demonstrate that it yields high-quality embeddings.
\item We evaluate \embedder and different state-of-the-art Euclidean embedding algorithms on a large set of graphs to see how heterogeneity affects the embedding quality.
\end{itemize}


\section{Embedding Method}

A graph $G = (V, E)$ can be perfectly reconstructed from an embedding if there is a threshold $\ell$ such that edges have length at most $\ell$ while non-adjacent vertices have larger distances.
This is formalized by the following \emph{threshold property}.
\begin{equation}
  \forall u,v\in V\colon \dist(u, v) \leq \ell \Leftrightarrow uv\in E
  \label{eq:threshold}
\end{equation}
Achieving this is a difficult task.
In fact, it is NP-hard (even $\exists \mathbb R$-complete) to determine whether a given graph can be embedded in the 2-dimensional Euclidean or hyperbolic plane \cite{DBLP:journals/comgeo/BreuK98, Compl_Some_Geomet_Topol_Probl-Schaef10, DBLP:journals/corr/abs-2301-05550}.
It is also NP-hard when using the dot product\footnote{For the dot product, higher value indicates stronger similarity, i.e., the inequality in \Cref{eq:threshold} has to be reversed.} \cite{DBLP:journals/dcg/KangM12}.
Nonetheless, we can use the threshold property as a guiding principle by minimizing a loss function $\Loss$ that achieves its minimum for embeddings that satisfying it.
In the following, we discuss our choice for $\Loss$, describe how we minimize it (mostly using gradient descent), and introduce a geometric data structure for the weighted space that speeds up the loss function evaluation.

\paragraph{Loss Function}

We note that directly interpreting \Cref{eq:threshold} as a minimization problem, i.e., trying to minimize the number of vertex pairs violating it, would yield a piecewise constant loss function, which makes using gradient descent infeasible.
Instead, we define a differentiable loss function $\Loss$ consisting of an attracting component $\Lattr$ that punishes too distance neighbors and a repelling component $\Lrep$ that punishes too close non-neighbors.
Formally, we define the loss $\Loss$ of an embedding as
\begin{equation}
  \Loss = \sum_{uv\in E} \Lattr(\dist(u, v)) + \sum_{uv\not\in E} \Lrep(\dist(u, v)).
  \label{eq:loss}
\end{equation}
We interpret the gradient of $\Lattr$ as attracting forces, which decreases the distance between neighboring nodes, while the gradient of $\Lrep$ is a repelling force between non-neighbors.
We call them $\fattr$ and $\frep$, respectively.

In this general framework, there are two degrees of freedom to fill: the distance functions $\dist$ and choices for the loss function $\Lattr$ and $\Lrep$.
The former is closely related to the embedding space and, as already discussed in the introduction, we use the weighted distances defined in \Cref{eq:weighted-distance}.

\begin{figure}[tb]
  \centering
  \includegraphics{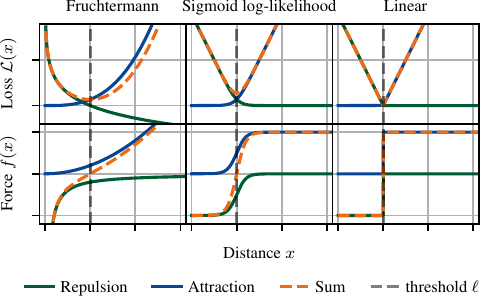}
  \caption{Different loss functions and resulting forces.
    Green lines correspond to $\Loss_{\text{rep}}$ and $\frep$, blue lines to $\Lattr$ and $\fattr$ and orange to their sum.
    The threshold $\ell$ is marked in gray.
  }
  \label{fig:loss}
\end{figure}

For the loss function, we considered various options that have been used in the literature; see \Cref{tab:loss} and \Cref{fig:loss}.
The minimum of $\Lattr(x) + \Lrep(x)$ corresponds to the threshold distance $\ell$ in \Cref{eq:threshold}.
Based on preliminary experiments, we decided for the \emph{linear loss function}.
Together with the very similar sigmoid log-likelihood, it showed the best performance.
Moreover, its simplicity has algorithmic advantages.
In particular, we can use the fact there are no forces between non-adjacent vertices with distance above the threshold $\ell$.

\begin{table*}[tb]
  \centering
  \caption{Different loss functions and their forces~\cite{DBLP:journals/spe/FruchtermanR91, DBLP:journals/tvcg/GansnerHN13, DBLP:conf/icdm/RahmanSA20}.
    The distance $\dist(u, v)$ is abbreviated with $x$ and $x'$ is its gradient with respect to the embedding of $u$. The indicator function $\mathbf{1}_{x>\ell}$ is $1$ if $x>\ell$ and $0$ otherwise.}
  \begin{tabular}{@{}lllll@{}}
    \toprule
    Loss function            & $\Lattr(x)$             & $\Loss_{\text{rep}}(x)$ & $\fattr(x)$                   & $\frep(x)$                     \\
    \midrule
    Fruchtermann \& Reingold & $x^3/3\ell $            & $-\ell^2\log(x)$        & $x'\cdot x^2/\ell$            & $-x'\cdot \ell^2/x$            \\
    Maxent Stress            & $(x-\ell)^2/\ell^2$     & $-\log(x)$              & $x'\cdot 2(x-\ell)/\ell^2$    & $-x'\cdot 1/x$                 \\
    Sigmoid log-likelihood   & $-\log(\sigma(\ell-x))$ & $-\log(\sigma(x-\ell))$ & $x'\cdot \sigma(x-\ell)$      & $-x'\cdot \sigma(\ell -x)$     \\
    Linear                   & $\max(x-\ell, 0)$       & $\max(\ell-x, 0)$       & $x'\cdot \mathbf{1}_{x>\ell}$ & $-x'\cdot \mathbf{1}_{x<\ell}$ \\
    \bottomrule
  \end{tabular}
  \label{tab:loss}
\end{table*}

\paragraph{Minimizing the Loss}

To every vertex $v \in V$, we need to assign a weight $w_v \in \mathbb R$ and a position $p_v \in \mathbb R^d$.
\embedder first fixes the weights using a simple degree-based estimation and then optimizes the positions using gradient descend.

We set the weights to $w_v = \deg(v)^{d / 8}$.
Without going too much into detail, the reasoning for this is roughly as follows.
Following \citet{DBLP:journals/tcs/BringmannKL19}, it makes sense to set $w_v = \deg(v)$ when embedding a graph with $d$-dimensional latent geometry into $d$-dimensional space.
However, for real-world networks, we do not know the true dimension, which leads to undesired effects in particular if the embedding dimension is too high.
The exponent $d / d'$ is a correction for embedding a graph that comes from $d'$ dimensions into $d$ dimensions.
Clearly $d'$ is a hyperparameter that depends on the input graph.
However, $d' = 8$ robustly lead to good results in our experiments.
This can be seen as an indication that the latent geometry underlying many real-world networks has quite low dimension.

After fixing the weights, \embedder finds positions by starting with a random initialization and then minimizing the linear loss function using batch gradient descent.
We apply Adam optimizer \cite{DBLP:journals/corr/KingmaB14} with an exponentially decreasing learning rate to update the embedding.
The optimization process ends, when the change in position is sufficiently small or a maximum number of iterations is reached.

To compute the gradient of $\Loss$ in every step, observe that the gradient is just the sum of attracting and repulsive forces contributed by every vertex pair.
We describe the gradient for the position $p_u$ of a fixed vertex $u$.
For every edge $uv \in E$ with $\dist(u, v) > \ell$, the derivative of $\dist(u, v)$ with respect to $p_u$ yields the vector $\frac{p_v - p_u}{\norm{p_u-p_v}} (w_uw_v)^{-1/d}$, which pulls $u$ towards $v$.
Similarly, for every non-edge $uv \notin E$ with $\dist(u, v) \le \ell$, we get the repulsive force $\frac{p_u - p_v}{\norm{p_u-p_v}} (w_uw_v)^{-1/d}$ pushing $u$ away from $v$.

Note that naively computing the gradient involves iterating over all $\Theta(n^2)$ vertex pairs, which is infeasible for larger graphs.
A common technique to reduce the time is negative sampling, i.e., consider all $m$ edges for the attractive forces and sample $\Oh(m)$ non-edges for the repelling forces.
\embedder does \emph{not} use negative sampling for two reasons.
First, it resulted in substantially worse weighted embeddings.
Secondly, as we aim for relatively low-dimensional embeddings, it is feasible to use a geometric data structure to find all non-neighbors at distance at most $\ell$.
As more distant non-neighbors are not relevant for the linear loss function, this lets us compute the gradient exactly.

\paragraph*{Geometric Data Structure}

For a given vertex $u$, we want to efficiently find all vertices with $\dist(u, v) \le \ell$, i.e., with $\norm{p_v-p_u} \leq \ell\cdot (w_uw_v)^{1/d}$.
A geometric data structure allowing queries of the type $\norm{p_v-p_u} \leq r$ with some radius $r$ is the R-tree \cite{R_Trees-Guttm84}.
Thus, if the weights were uniform, one could simply use an R-tree to find all relevant non-neighbors of $u$.
However, the radius $r$ depends on $w_v$, which requires querying different radii for vertices of different weight.
To resolve this, we use a similar technique that has been used to generate graphs \cite{DBLP:journals/tcs/BringmannKL19, Blaesius_Friedrich_Katzmann_Meyer_Penschuck_Weyand_2022}.
We partition the nodes into classes with similar weights, creating one R-tree for each class, and query the different R-trees with different radii.
More specifically, let $V_i=\set{v\in V\mid 2^{i-1}\leq w_v < 2^{i}}$.
We create one R-tree $T_i$ for each non-empty weight class $V_i$.
Querying $T_i$ with radius $r_i(u) = \ell \cdot (w_u2^{i})^{1/d}$ yields all vertices $v \in V_i$ with weighted distance at most $\ell$ to $u$, plus some false positives with weighted distance up to $2\ell$.
Thus, the union over all R-trees gives a superset of the nodes with weighted distance $\ell$ to $u$, which includes all non-neighbors relevant for repulsive forces.


\section{Experiments}

We evaluate the performance of \embedder in comparison to different other embedding methods.
Additionally we investigate the impact of a graph's heterogeneity on the embedding quality and on the relevance of using a weighted space.

\subsection{Setup}

We compare \embedder with the state-of-the-art Euclidean embedders \deepwalkemb~\cite{DBLP:conf/kdd/PerozziAS14}, \verseemb~\cite{DBLP:conf/www/TsitsulinMKM18}, and \ftovemb~\cite{DBLP:conf/icdm/RahmanSA20}, as well as with the hyperbolic embedder \poincareemb \cite{DBLP:conf/nips/NickelK17}.
We used the default parameters for \deepwalkemb, sigmoid and negative sampling (option 6) for \ftovemb, and adjacency similarity (verse\textunderscore{}neigh) for \verseemb.
The default parameters for \poincareemb did not yield good results and we used $\text{lr} = 0.3$, $\text{epochs} = 300$, and $\text{batchsize} = 50$.
All experiments were run on a machine with two AMD EPYC Zen2 7742 (64 cores) with 2.25-3.35 GHz frequency, 1024GB RAM and 256MB L3 cache.
To account for randomness, each data point is the average of five runs.
To get comparable running times, computation is single-threaded.
The only exception is the much slower \poincareemb, for which we compute only one embedding and use all 128 available cores.

\subsubsection{Data}

We use a set of real-world networks described by \citet{DBLP:journals/talg/BlasiusF24} restricted to graphs with $n \le \SI{10}{k}$ nodes and $m \le \SI{50}{k}$ edges, which amounts to \num{2173} graphs.
We refer to this data set as \texttt{real}.
Moreover, we explicitly feature a set of \num{13} selected networks.
This set is not disjoint to \texttt{real} but additionally contains some commonly considered larger networks.
We also consider randomly generated graphs, which are described later.
Each graph is considered as unlabeled, undirected, and unweighted, and is reduced to its largest connected component before embedding.

\subsubsection{Graph Reconstruction}

We evaluate the embedding quality based on reconstruction accuracy.
To make this precise, let $t$ be some threshold and let $E(t)$ be the set of \emph{predicted edges} with respect to $t$, i.e., all pairs of nodes with embedding distance at most $t$.
The \emph{precision} $\Prec(t)\coloneqq \frac{\abs{E\cap E(t)}}{\abs{E(t)}}$ is the fraction of predicted edges that are actual edges.
The \emph{recall} $\Rec(t)\coloneqq \frac{\abs{E\cap E(t)}}{\abs{E}}$ is the fraction of actual edges that are predicted.
The \emph{F1-score} is the harmonic mean of precision and recall where $t$ is chosen to maximize the score.
For large graphs we compute an approximated F1-score by sampling $10\cdot m$ non-edges.
We note that we also considered mean average precision as well as link prediction.
We do not report these scores here as they do not provide additional  insights.

\subsection{Evaluation}

\begin{figure}[t]
	\centering
	\includegraphics{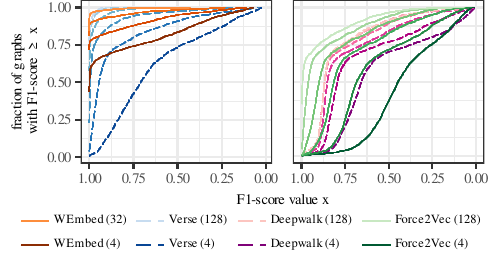}
	\caption{Embedding quality for the \texttt{real} data set.
          Each line represents the CDF of the F1-score (note the reversed $x$-axis) for an embedder in a fixed dimension.
          Dimensions for the different embedders increase in powers of $2$: $4, 8, 16, 32$ for \embedder and additionally $64, 128$ for the others.}
	\label{fig:quality_cactus}
\end{figure}

\begin{table*}[tb]
	\centering
	\caption{
	F1-score and embedding time (gray) in seconds on selected real-world graphs.
	We report results for multiple different dimensions.
	We ran experiments with \poincareemb in parallel and mark these running times with an $\ast$.
	}
	\label{tab:real_world_results}

\definecolor{myGray}{rgb}{1,1,1}
\definecolor{myGray2}{rgb}{0.5,0.5,0.5}
\newcommand{\tablenumber}[1]{\small\ensuremath{#1}}
\newcommand{\tabletime}[1]{\footnotesize\textcolor{myGray2}{\ensuremath{\num{#1}}}}
\newcommand{\size}[2]{\footnotesize\ensuremath{\abs{V}=\num{#1},\ \abs{E}=\num{#2}}}
\newcommand{\myast}{\footnotesize\textcolor{myGray2}{$^\ast$}}
\newcommand{\mydash}{\small\textcolor{myGray2}{$-$}}
\newcommand{\graphname}[1]{\small\texttt{#1}}

\begin{tabular}{@{}lclllcllllcl@{}}
    \toprule
                                                              &  & \multicolumn{3}{c}{\embedder} &                     & \multicolumn{4}{c}{\verseemb} &  & \multirow{2}{1.58cm}{\textsc{Poincaré} \textsc{Embed} (8)}                                                                                                \\
    \cmidrule(lr){3-5} \cmidrule(lr){7-10}
    Graph                                                     &  & \tablenumber{8}               & \tablenumber{16}    & \tablenumber{32}              &  & \tablenumber{8}                                             & \tablenumber{16}    & \tablenumber{32}    & \tablenumber{128}   &  &                         \\
    \midrule

    \cellcolor{white}\graphname{AstroPh1}                     &  & \tablenumber{0.776}           & \tablenumber{0.969} & \tablenumber{1}               &  & \tablenumber{0.425}                                         & \tablenumber{0.804} & \tablenumber{0.955} & \tablenumber{0.999} &  & \tablenumber{-}         \\
    \size{14845}{119652}                                      &  & \tabletime{610}               & \tabletime{2080}    & \tabletime{3644}              &  & \tabletime{172}                                             & \tabletime{200}     & \tabletime{266}     & \tabletime{908}     &  & \mydash                 \\
    \rowcolor{myGray}\cellcolor{white}\graphname{AstroPh2}    &  & \tablenumber{0.619}           & \tablenumber{0.836} & \tablenumber{0.997}           &  & \tablenumber{0.278}                                         & \tablenumber{0.609} & \tablenumber{0.868} & \tablenumber{0.997} &  & \tablenumber{0.565}     \\
    \size{17903}{196972}                                      &  & \tabletime{989}               & \tabletime{3678}    & \tabletime{5907}              &  & \tabletime{217}                                             & \tabletime{252}     & \tabletime{338}     & \tabletime{1741}    &  & \tabletime{38848}\myast \\
    \rowcolor{myGray}\cellcolor{white}\graphname{BlogCatalog} &  & \tablenumber{0.309}           & \tablenumber{0.389} & \tablenumber{0.467}           &  & \tablenumber{0.335}                                         & \tablenumber{0.359} & \tablenumber{0.435} & \tablenumber{0.834} &  & \tablenumber{0.408}     \\
    \size{10312}{333983}                                      &  & \tabletime{660}               & \tabletime{1517}    & \tabletime{3408}              &  & \tabletime{138}                                             & \tabletime{164}     & \tabletime{220}     & \tabletime{428}     &  & \tabletime{64326}\myast \\
    \rowcolor{myGray}\cellcolor{white}\graphname{CAIDA}       &  & \tablenumber{0.737}           & \tablenumber{0.981} & \tablenumber{0.999}           &  & \tablenumber{0.319}                                         & \tablenumber{0.721} & \tablenumber{0.899} & \tablenumber{0.994} &  & \tablenumber{0.458}     \\
    \size{26475}{53381}                                       &  & \tabletime{897}               & \tabletime{3393}    & \tabletime{9667}              &  & \tabletime{276}                                             & \tabletime{316}     & \tabletime{418}     & \tabletime{2910}    &  & \tabletime{11116}\myast \\
    \rowcolor{myGray}\cellcolor{white}\graphname{Citeseer}    &  & \tablenumber{0.999}           & \tablenumber{1}     & \tablenumber{1}               &  & \tablenumber{0.641}                                         & \tablenumber{0.924} & \tablenumber{0.977} & \tablenumber{0.993} &  & \tablenumber{0.669}     \\
    \size{2110}{3668}                                         &  & \tabletime{13}                & \tabletime{26}      & \tabletime{61}                &  & \tabletime{24}                                              & \tabletime{26}      & \tabletime{31}      & \tabletime{65}      &  & \tabletime{569}\myast   \\
    \rowcolor{myGray}\cellcolor{white}\graphname{CondMat}     &  & \tablenumber{0.736}           & \tablenumber{0.989} & \tablenumber{1}               &  & \tablenumber{0.275}                                         & \tablenumber{0.758} & \tablenumber{0.946} & \tablenumber{0.999} &  & \tablenumber{-}         \\
    \size{21363}{91286}                                       &  & \tabletime{971}               & \tabletime{4065}    & \tabletime{7498}              &  & \tabletime{251}                                             & \tabletime{283}     & \tabletime{362}     & \tabletime{2239}    &  & \mydash                 \\
    \rowcolor{myGray}\cellcolor{white}\graphname{Cora}        &  & \tablenumber{1}               & \tablenumber{1}     & \tablenumber{1}               &  & \tablenumber{0.646}                                         & \tablenumber{0.930} & \tablenumber{0.986} & \tablenumber{1}     &  & \tablenumber{0.676}     \\
    \size{2485}{5069}                                         &  & \tabletime{17}                & \tabletime{40}      & \tabletime{100}               &  & \tabletime{28}                                              & \tabletime{30}      & \tabletime{36}      & \tabletime{76}      &  & \tabletime{813}\myast   \\
    \rowcolor{myGray}\cellcolor{white}\graphname{DBLP}        &  & \tablenumber{0.562}           & \tablenumber{0.857} & \tablenumber{1}               &  & \tablenumber{0.255}                                         & \tablenumber{0.564} & \tablenumber{0.844} & \tablenumber{0.989} &  & \tablenumber{-}         \\
    \size{12495}{49563}                                       &  & \tabletime{358}               & \tabletime{1410}    & \tabletime{2497}              &  & \tabletime{133}                                             & \tabletime{153}     & \tabletime{198}     & \tabletime{474}     &  & \mydash                 \\
    \rowcolor{myGray}\cellcolor{white}\graphname{GrQc}        &  & \tablenumber{0.981}           & \tablenumber{1}     & \tablenumber{1}               &  & \tablenumber{0.689}                                         & \tablenumber{0.965} & \tablenumber{0.999} & \tablenumber{1}     &  & \tablenumber{-}         \\
    \size{4158}{13422}                                        &  & \tabletime{50}                & \tabletime{98}      & \tabletime{254}               &  & \tabletime{44}                                              & \tabletime{50}      & \tabletime{61}      & \tabletime{130}     &  & \mydash                 \\
    \rowcolor{myGray}\cellcolor{white}\graphname{HepPh}       &  & \tablenumber{0.869}           & \tablenumber{0.989} & \tablenumber{1}               &  & \tablenumber{0.516}                                         & \tablenumber{0.840} & \tablenumber{0.957} & \tablenumber{0.999} &  & \tablenumber{0.757}     \\
    \size{11203}{117618}                                      &  & \tabletime{376}               & \tabletime{1064}    & \tabletime{2065}              &  & \tabletime{127}                                             & \tabletime{146}     & \tabletime{188}     & \tabletime{402}     &  & \tabletime{22384}\myast \\
    \rowcolor{myGray}\cellcolor{white}\graphname{Internet}    &  & \tablenumber{0.724}           & \tablenumber{0.962} & \tablenumber{1}               &  & \tablenumber{0.334}                                         & \tablenumber{0.718} & \tablenumber{0.899} & \tablenumber{0.990} &  & \tablenumber{0.520}     \\
    \size{23748}{58414}                                       &  & \tabletime{821}               & \tabletime{2839}    & \tabletime{6929}              &  & \tabletime{248}                                             & \tabletime{280}     & \tabletime{375}     & \tabletime{2836}    &  & \tabletime{12174}\myast \\
    \rowcolor{myGray}\cellcolor{white}\graphname{Pubmed}      &  & \tablenumber{0.705}           & \tablenumber{0.989} & \tablenumber{1}               &  & \tablenumber{0.217}                                         & \tablenumber{0.637} & \tablenumber{0.907} & \tablenumber{0.994} &  & \tablenumber{0.422}     \\
    \size{19717}{44324}                                       &  & \tabletime{601}               & \tabletime{2420}    & \tabletime{5657}              &  & \tabletime{216}                                             & \tabletime{245}     & \tabletime{310}     & \tabletime{1745}    &  & \tabletime{9167}\myast  \\
    \rowcolor{myGray}\cellcolor{white}\graphname{RouteViews}  &  & \tablenumber{0.905}           & \tablenumber{1}     & \tablenumber{1}               &  & \tablenumber{0.537}                                         & \tablenumber{0.887} & \tablenumber{0.985} & \tablenumber{1}     &  & \tablenumber{0.542}     \\
    \size{6474}{12572}                                        &  & \tabletime{100}               & \tabletime{232}     & \tabletime{590}               &  & \tabletime{62}                                              & \tabletime{70}      & \tabletime{92}      & \tabletime{199}     &  & \tabletime{2624}\myast  \\
    \bottomrule
\end{tabular}

\end{table*}

The embedding quality on the \texttt{real} data set is shown as the cumulative distribution function (CDF) of the F1-scores in \Cref{fig:quality_cactus}.
A line going through the point $(x, y)$ indicates that a fraction of $y$ instances has an F1-score of $x$ or higher.
For example, \embedder~(8) yields an F1-score very close to \num{1} for over \SI{75}{\%} of graphs.
We can also observe the trend that a higher number of dimensions allows for better accuracy.

Comparing the embedders, \verseemb clearly has the best accuracy among the Euclidean approaches.
In the comparison to \embedder, note that \embedder uses at most 32 dimensions while we show results for \verseemb using up to 128 dimensions.
Comparing equal dimensions, \embedder clearly dominates \verseemb.
Moreover, even \embedder~(32) is comparable with \verseemb~(128), with \embedder~(32) getting very high F1-scores for more graphs, while \verseemb~(128) excels for lower scores, i.e., on graphs that are difficult to embed.

Table~\ref{tab:real_world_results} reports detailed results for \num{13} selected graph.
The comparison with \verseemb matches the previous observation: \embedder outperforms \verseemb in low dimensions, sometimes by a lot.
Moreover, while \verseemb excels for \num{128} dimensions on the difficult-to-embed graph \texttt{BlogCatalog}, \embedder often achieves better results, even in fewer dimensions.

The hyperbolic embedder \poincareemb yields better accuracy than \verseemb for $d = 8$, confirming the superiority of weighted and hyperbolic over Euclidean geometry.
However, the comparison between \poincareemb and \embedder goes clearly in favor of \embedder, with \texttt{BlogCatalog} being the only exception.
Additionally, \poincareemb is much slower; usually by multiple orders of magnitude, despite using \num{128} cores.
This worse performance is also the reason why we limited the \poincareemb experiments to \num{8} dimensions and a smaller selection of graphs.

The running time of \verseemb is usually better then that of \embedder, though the exact comparison depends on the graph.
For low dimensions, the running times are in the same order of magnitude.
Moreover, we note that a fair comparison is probably to compare the $d = 32$ for \embedder with $d = 128$ for \verseemb, where the embeddings have similar quality.

\begin{figure*}[ht]
	\centering
	\includegraphics{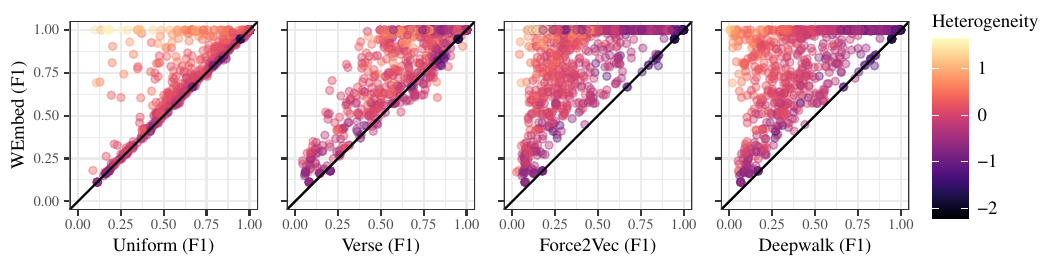}
	\caption{Comparing the quality for $8$-dimensional embeddings for the \texttt{real} data set.
          Each point is one graph with the F1-score achieved by \embedder on the $y$-axis and the F1-score by Euclidean approaches on the $x$-axis.
          The color indicates the heterogeneity of the degree distribution.  Euclidean (first column) is \embedder but with uniform weight.}
	\label{fig:real_world_graphs}
\end{figure*}

\paragraph{The Effect of Heterogeneity}

\begin{figure*}[t]
	\centering
	\includegraphics{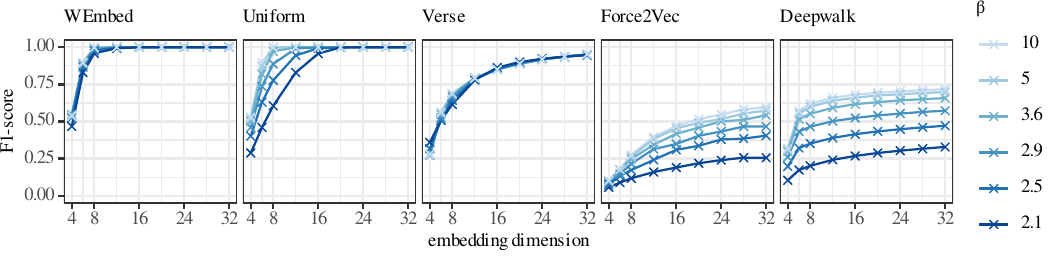}
	\caption{Embedding quality for GIRGs with different power-law exponents.
          Euclidean refers to \embedder with uniform weights.}
	\label{fig:girg_ples}
\end{figure*}

Here we assess the assumption that weighted embeddings are better suited for graphs with heterogeneous degree distribution.
Figure~\ref{fig:real_world_graphs} shows a scatter plot comparing \embedder to Euclidean approaches, with color indicating the \emph{heterogeneity} of the graph, which is defined as the $\log_{10}$ of the coefficient of variation of the degree distribution \cite{DBLP:journals/talg/BlasiusF24}.
Comparing \embedder to itself but with uniform weights (Figure~\ref{fig:real_world_graphs} left) shows a clear trend that graphs with higher heterogeneity benefit more strongly from using weights.
The same trend can be seen in comparison to \ftovemb and \deepwalkemb.  The effect is also present for \verseemb but less pronounced.

As the real-world networks differ in more than just their heterogeneity, we additionally use generated graphs that let us control the heterogeneity while keeping everything else fixed.
We consider \emph{geometric inhomogeneous random graphs} (GIRGs) \cite{DBLP:journals/tcs/BringmannKL19}, using the generator by \cite{Blaesius_Friedrich_Katzmann_Meyer_Penschuck_Weyand_2022}.
We generate GIRGs with $n = \SI{10}{k}$ nodes, average degree \num{15}, temperature \num{0.1}, and dimension 4.
Additionally we vary the power-law exponent $\beta$, which controls the heterogeneity of the degree distribution (high heterogeneity close to $\beta = 2$ and almost uniform for $\beta = 10$).
Figure~\ref{fig:girg_ples} shows the embedding quality of the different embedders depending on the embedding dimension and the power-law exponent.
One can clearly see the power-law exponent does not make a difference for \embedder, while its variant with uniform weights, as well as \ftovemb and \deepwalkemb have more difficulties with lower $\beta$ (i.e., higher heterogeneity).
Interestingly, the quality for \verseemb is independent of $\beta$.
We mention potential reasons for this when discussing the target space in the next section.


\section{Discussion and Conclusion}

Our experiments clearly demonstrate that \embedder is superior to other state-of-the-art embedding algorithms for low dimensions.
With this, \embedder represents two types of contributions.
First, it is a new tool for efficiently computing low-dimensional representations of high quality.
Secondly, it serves as a proof-of-concept to understand the usefulness of weighted embeddings.
That being said, we believe that both perspective, i.e., engineering new tools as well as deepening our understanding of embeddings, still leave room for improvement.
In the following, we discuss which future directions we find promising in this regard.
Moreover, we address further minor insights we gained while building \embedder.

\paragraph{Target Space for Heterogeneous Graphs}

Despite the close connection between hyperbolic geometry and Euclidean space with weights \cite{DBLP:journals/tcs/BringmannKL19}, there are clearly some differences, e.g., the weighted distance in Equation~\eqref{eq:weighted-distance} is not actually a metric as it does not satisfy the triangular inequality.
The impact of these differences on embeddings is not yet fully understood.
One difference becoming clear with this paper is that using weighted rather than hyperbolic spaces makes the embedding problem algorithmically more tractable.
This raises the question whether the gained simplicity comes at the cost of representative power.
Our experiments clearly agree with the insight from network science that weighted space is as suited to represent heterogeneous graphs as hyperbolic geometry.
We conjecture that this equivalence is also given for hierarchical structures like trees, even if their degree distribution is uniform.
We note that \embedder in its current state would struggle to embed trees with uniform degree distribution, as it would assign every vertex the same weight yielding just a Euclidean embedding.
One way to mitigate this is to take the transitive closure of the tree.
Interestingly, this technique has also been used for hyperbolic embeddings by \citet{DBLP:conf/nips/NickelK17}.
Thus, struggling to embed uniform hierarchical structures is not unique to \embedder and we believe that the reason is purely algorithmic and not due to the target space.
It is an interesting direction for future research to rigorously study these differences between hyperbolic and weighted space and how they can represent different structures.

Another observation related to the target space is that \verseemb is not heavily impacted by the heterogeneity.
\verseemb uses the dot product as similarity measure.
For normalized vectors, the dot product just measures the angle between vectors and thus there is a close connection to distances on a sphere, which behaves close to Euclidean.
However, \verseemb is not restricted to normalized vectors, and thus the similarity measure is scaled by the length of the vectors.
We believe that this is the reason that \verseemb can handle heterogeneous graphs to some degree.
It seems fruitful to better understand how the dot product compares to hyperbolic or weighted space and to further explore related spaces.

\paragraph{Algorithmic Techniques}

Although \embedder already produces high quality embeddings and is reasonably efficient, there is still room for improvements.
The most obvious lies in optimizing the weights, which are currently chosen with a simple heuristic solely based on the node degrees.
We note, however, that optimizing the weights is not trivial.
Preliminary experiments indicated that using gradient descend does not work well for the weights.
Nonetheless, we believe that there is quite some potential using other techniques.
In particular, we would be interested to see whether using other centrality measures than just the node degree yields good weights also for hierarchical structures with uniform degrees.

Another approach for improving the embedding quality commonly used for visualization (i.e., very low dimensional embeddings) is a multi-level approach~\cite{Drawin_Large_Graph_Poten_Field-HachulJuenger05,Drawin_Large_Graph_Multil_Maxen_Stres_Optim-Meyer18}.
The core idea is to hierarchically group the nodes into clusters and embed the rough structure by finding positions for the clusters before placing individual nodes.
This helps to untangle the graph.
Preliminary experiments with \embedder showed that this only helps in very low dimensions.
This makes sense as, intuitively, nodes are less likely to get trapped in local optima as there are more directions to get around an obstacle.
Thus, the gradient becomes more well-behaved in higher dimensions, eliminating the need for a multi-level approach.

Finally, we believe that exploring more advanced negative sampling techniques as well as improving the geometric data structure has the potential to improve the running time.


\paragraph{Applications}

We demonstrated that \embedder can produce very accurate low-dimensional representation of graphs by embedding them into a weighted space.
An obvious next question is how the insight that weighted space is well suited to host heterogeneous data, as well as the tool \embedder itself, can be utilized for various applications.
This involves down-stream tasks on graphs like node classification, the extension to, e.g., graph neural networks, and the applicability to other domains like word embeddings for natural language processing or computer vision, where hyperbolic embeddings have been observed to be beneficial.
We believe that weighted embeddings are simpler to adapt and that the resulting models are easier to optimize, while providing the same benefits that hyperbolic geometry promises.


\bibliography{hierarchical-grig-embedder}

\begin{thebibliography}{47}
\providecommand{\natexlab}[1]{#1}

\bibitem[{Bieker et~al.(2023)Bieker, Bl{\"{a}}sius, Dohse, and
  Jungeblut}]{DBLP:journals/corr/abs-2301-05550}
Bieker, N.; Bl{\"{a}}sius, T.; Dohse, E.; and Jungeblut, P. 2023.
\newblock Recognizing Unit Disk Graphs in Hyperbolic Geometry is
  {\(\exists\mathbb R\)}-Complete.
\newblock \emph{CoRR}, abs/2301.05550.

\bibitem[{Bl{\"{a}}sius and Fischbeck(2024)}]{DBLP:journals/talg/BlasiusF24}
Bl{\"{a}}sius, T.; and Fischbeck, P. 2024.
\newblock On the External Validity of Average-case Analyses of Graph
  Algorithms.
\newblock \emph{{ACM} Trans. Algorithms}, 20(1): 10:1--10:42.

\bibitem[{Bläsius, Friedrich, and
  Katzmann(2021)}]{Force_Direc_Embed_Scale_Free-Blasius21}
Bläsius, T.; Friedrich, T.; and Katzmann, M. 2021.
\newblock Force-Directed Embedding of Scale-Free Networks in the Hyperbolic
  Plane.
\newblock In \emph{Symposium on Experimental Algorithms (SEA)}, Leibniz
  International Proceedings in Informatics (LIPIcs), 22:1--22:18. Schloss
  Dagstuhl -- Leibniz-Zentrum f{\"u}r Informatik.

\bibitem[{Bläsius et~al.(2022)Bläsius, Friedrich, Katzmann, Meyer, Penschuck,
  and Weyand}]{Blaesius_Friedrich_Katzmann_Meyer_Penschuck_Weyand_2022}
Bläsius, T.; Friedrich, T.; Katzmann, M.; Meyer, U.; Penschuck, M.; and
  Weyand, C. 2022.
\newblock Efficiently Generating Geometric Inhomogeneous and Hyperbolic Random
  Graphs.
\newblock \emph{Network Science}, 10(4): 361--380.

\bibitem[{Bläsius et~al.(2016)Bläsius, Friedrich, Krohmer, and
  Laue}]{Effic_Embed_Scale_Graph_ESA2016}
Bläsius, T.; Friedrich, T.; Krohmer, A.; and Laue, S. 2016.
\newblock Efficient Embedding of Scale-Free Graphs in the Hyperbolic Plane.
\newblock In \emph{European Symposium on Algorithms (ESA)}, 16:1--16:18.

\bibitem[{Boguñá, Papadopoulos, and Krioukov(2010)}]{Sustain_Inter-Bogun10}
Boguñá, M.; Papadopoulos, F.; and Krioukov, D. 2010.
\newblock Sustaining the Internet with Hyperbolic Mapping.
\newblock \emph{Nature Communications}, 1(62).

\bibitem[{Breu and Kirkpatrick(1998)}]{DBLP:journals/comgeo/BreuK98}
Breu, H.; and Kirkpatrick, D.~G. 1998.
\newblock Unit disk graph recognition is NP-hard.
\newblock \emph{Comput. Geom.}, 9(1-2): 3--24.

\bibitem[{Bringmann, Keusch, and
  Lengler(2019)}]{DBLP:journals/tcs/BringmannKL19}
Bringmann, K.; Keusch, R.; and Lengler, J. 2019.
\newblock Geometric Inhomogeneous Random Graphs.
\newblock \emph{Theor. Comput. Sci.}, 760: 35--54.

\bibitem[{Chamberlain, Clough, and
  Deisenroth(2017)}]{Neural_Embed_Graph_Hyper_Space-Chamb17}
Chamberlain, B.~P.; Clough, J.~R.; and Deisenroth, M.~P. 2017.
\newblock Neural Embeddings of Graphs in Hyperbolic Space.
\newblock \emph{CoRR}, abs/1705.10359.

\bibitem[{Chami et~al.(2019)Chami, Ying, R{\'{e}}, and
  Leskovec}]{Hyper_Graph_Convol_Neural_Networ-Chami19}
Chami, I.; Ying, Z.; R{\'{e}}, C.; and Leskovec, J. 2019.
\newblock Hyperbolic Graph Convolutional Neural Networks.
\newblock In \emph{Advances in Neural Information Processing Systems
  (NeurIPS)}, 4869--4880.

\bibitem[{Choudhary et~al.(2022)Choudhary, Rao, Subbian, Sengamedu, and
  Reddy}]{Hyper_Neural_Networ-Choud22}
Choudhary, N.; Rao, N.; Subbian, K.; Sengamedu, S.~H.; and Reddy, C.~K. 2022.
\newblock Hyperbolic Neural Networks: Theory, Architectures and Applications.
\newblock In \emph{International Conference on Knowledge Discovery and Data
  Mining (KDD)}, 4778--4779. Association for Computing Machinery.

\bibitem[{Fruchterman and Reingold(1991)}]{DBLP:journals/spe/FruchtermanR91}
Fruchterman, T. M.~J.; and Reingold, E.~M. 1991.
\newblock Graph Drawing by Force-directed Placement.
\newblock \emph{Softw. Pract. Exp.}, 21(11): 1129--1164.

\bibitem[{Ganea, B{\'{e}}cigneul, and
  Hofmann(2018)}]{Hyper_Entail_Cones_Learn_Hierar_Embed-Ganea18}
Ganea, O.; B{\'{e}}cigneul, G.; and Hofmann, T. 2018.
\newblock Hyperbolic Entailment Cones for Learning Hierarchical Embeddings.
\newblock In \emph{International Conference on Machine Learning (ICML)},
  volume~80 of \emph{Proceedings of Machine Learning Research}, 1632--1641.
  {PMLR}.

\bibitem[{Ganea, Becigneul, and Hofmann(2018)}]{Hyper_Neural_Networ-Ganea18}
Ganea, O.; Becigneul, G.; and Hofmann, T. 2018.
\newblock Hyperbolic Neural Networks.
\newblock In \emph{Advances in Neural Information Processing Systems
  (NeurIPS)}, volume~31. Curran Associates, Inc.

\bibitem[{Gansner, Hu, and North(2013)}]{DBLP:journals/tvcg/GansnerHN13}
Gansner, E.~R.; Hu, Y.; and North, S.~C. 2013.
\newblock A Maxent-Stress Model for Graph Layout.
\newblock \emph{{IEEE} Trans. Vis. Comput. Graph.}, 19(6): 927--940.

\bibitem[{Guttman(1984)}]{R_Trees-Guttm84}
Guttman, A. 1984.
\newblock R-Trees: {A} Dynamic Index Structure for Spatial Searching.
\newblock In \emph{{SIGMOD} International Conference on Management of Data
  (SIGMOD)}, 47--57. {ACM} Press.

\bibitem[{Hachul and
  Jünger(2005)}]{Drawin_Large_Graph_Poten_Field-HachulJuenger05}
Hachul, S.; and Jünger, M. 2005.
\newblock Drawing Large Graphs with a Potential-Field-Based Multilevel
  Algorithm.
\newblock In \emph{Graph Drawing}, 285--295. Springer Berlin Heidelberg.

\bibitem[{Kang and M{\"{u}}ller(2012)}]{DBLP:journals/dcg/KangM12}
Kang, R.~J.; and M{\"{u}}ller, T. 2012.
\newblock Sphere and Dot Product Representations of Graphs.
\newblock \emph{Discret. Comput. Geom.}, 47(3): 548--568.

\bibitem[{Khrulkov et~al.(2020)Khrulkov, Mirvakhabova, Ustinova, Oseledets, and
  Lempitsky}]{Hyper_Image_Embed-Khrul20}
Khrulkov, V.; Mirvakhabova, L.; Ustinova, E.; Oseledets, I.~V.; and Lempitsky,
  V.~S. 2020.
\newblock Hyperbolic Image Embeddings.
\newblock In \emph{Conference on Computer Vision and Pattern Recognition
  (CVPR)}, 6417--6427. Computer Vision Foundation / {IEEE}.

\bibitem[{Kingma and Ba(2015)}]{DBLP:journals/corr/KingmaB14}
Kingma, D.~P.; and Ba, J. 2015.
\newblock Adam: {A} Method for Stochastic Optimization.
\newblock In \emph{International Conference on Learning Representations
  (ICLR)}.

\bibitem[{Kisfaludi-Bak and van
  Wordragen(2024)}]{Quadt_Stein_Spann_Approx_Neares-KisfalWordr24}
Kisfaludi-Bak, S.; and van Wordragen, G. 2024.
\newblock A Quadtree, a Steiner Spanner, and Approximate Nearest Neighbours in
  Hyperbolic Space.
\newblock In \emph{International Symposium on Computational Geometry (SoCG)},
  volume 293 of \emph{Leibniz International Proceedings in Informatics
  (LIPIcs)}, 68:1--68:15. Schloss Dagstuhl -- Leibniz-Zentrum f{\"u}r
  Informatik.

\bibitem[{Krioukov et~al.(2010)Krioukov, Papadopoulos, Kitsak, Vahdat, and
  Bogu\~n\'a}]{Hyper_Geomet_Compl_Networ-Kriouk10}
Krioukov, D.; Papadopoulos, F.; Kitsak, M.; Vahdat, A.; and Bogu\~n\'a, M.
  2010.
\newblock Hyperbolic Geometry of Complex Networks.
\newblock \emph{Phys. Rev. E}, 82: 036106.

\bibitem[{Law et~al.(2019)Law, Liao, Snell, and
  Zemel}]{Loren_Distan_Learn_Hyper_Repres-Law19}
Law, M.~T.; Liao, R.; Snell, J.; and Zemel, R.~S. 2019.
\newblock Lorentzian Distance Learning for Hyperbolic Representations.
\newblock In \emph{International Conference on Machine Learning (ICML)},
  volume~97 of \emph{Proceedings of Machine Learning Research}, 3672--3681.
  {PMLR}.

\bibitem[{Liu, Nickel, and Kiela(2019)}]{Hyper_Graph_Neural_Networ-Liu19}
Liu, Q.; Nickel, M.; and Kiela, D. 2019.
\newblock Hyperbolic Graph Neural Networks.
\newblock In \emph{Advances in Neural Information Processing Systems
  (NeurIPS)}, 8228--8239.

\bibitem[{Meyerhenke, Nöllenburg, and
  Schulz(2018)}]{Drawin_Large_Graph_Multil_Maxen_Stres_Optim-Meyer18}
Meyerhenke, H.; Nöllenburg, M.; and Schulz, C. 2018.
\newblock Drawing Large Graphs by Multilevel Maxent-Stress Optimization.
\newblock \emph{{IEEE} Trans. Vis. Comput. Graph.}, 24(5): 1814--1827.

\bibitem[{Mishne et~al.(2023)Mishne, Wan, Wang, and
  Yang}]{Numer_Stabil_Hyper_Repres_Learn-Mishn23}
Mishne, G.; Wan, Z.; Wang, Y.; and Yang, S. 2023.
\newblock The Numerical Stability of Hyperbolic Representation Learning.
\newblock In \emph{International Conference on Machine Learning (ICML)}, volume
  202 of \emph{Proceedings of Machine Learning Research}, 24925--24949. PMLR.

\bibitem[{Nickel and Kiela(2017)}]{DBLP:conf/nips/NickelK17}
Nickel, M.; and Kiela, D. 2017.
\newblock Poincar{\'{e}} Embeddings for Learning Hierarchical Representations.
\newblock In \emph{Advances in Neural Information Processing Systems
  (NeurIPS)}, 6338--6347.

\bibitem[{Nickel and Kiela(2018)}]{DBLP:conf/icml/NickelK18}
Nickel, M.; and Kiela, D. 2018.
\newblock Learning Continuous Hierarchies in the Lorentz Model of Hyperbolic
  Geometry.
\newblock In \emph{International Conference on Machine Learning (ICML)},
  volume~80 of \emph{Proceedings of Machine Learning Research}, 3776--3785.
  {PMLR}.

\bibitem[{Papadopoulos, Aldecoa, and
  Krioukov(2015)}]{Networ_Geomet_Infer_Using_Common_Neigh-Papad15}
Papadopoulos, F.; Aldecoa, R.; and Krioukov, D. 2015.
\newblock Network Geometry Inference Using Common Neighbors.
\newblock \emph{Phys. Rev. E}, 92: 022807.

\bibitem[{Papadopoulos, Psomas, and
  Krioukov(2015)}]{Networ_Mappin_Replay_Hyper_Growt-Papad15}
Papadopoulos, F.; Psomas, C.; and Krioukov, D. 2015.
\newblock Network Mapping by Replaying Hyperbolic Growth.
\newblock \emph{IEEE/ACM Transactions on Networking}, 23(1): 198--211.

\bibitem[{Peng et~al.(2022)Peng, Varanka, Mostafa, Shi, and
  Zhao}]{Hyper_Deep_Neural_Networ-Peng22}
Peng, W.; Varanka, T.; Mostafa, A.; Shi, H.; and Zhao, G. 2022.
\newblock Hyperbolic Deep Neural Networks: A Survey.
\newblock \emph{IEEE Transactions on Pattern Analysis and Machine
  Intelligence}, 44(12): 10023--10044.

\bibitem[{Perozzi, Al{-}Rfou, and Skiena(2014)}]{DBLP:conf/kdd/PerozziAS14}
Perozzi, B.; Al{-}Rfou, R.; and Skiena, S. 2014.
\newblock DeepWalk: online learning of social representations.
\newblock In \emph{International Conference on Knowledge Discovery and Data
  Mining (KDD)}, 701--710. {ACM}.

\bibitem[{Rahman, Sujon, and Azad(2020)}]{DBLP:conf/icdm/RahmanSA20}
Rahman, M.~K.; Sujon, M.~H.; and Azad, A. 2020.
\newblock {Force2Vec}: Parallel Force-Directed Graph Embedding.
\newblock In \emph{International Conference on Data Mining (ICDM)}, 442--451.
  {IEEE}.

\bibitem[{Sala et~al.(2018)Sala, Sa, Gu, and
  R{\'{e}}}]{DBLP:conf/icml/SalaSGR18}
Sala, F.; Sa, C.~D.; Gu, A.; and R{\'{e}}, C. 2018.
\newblock Representation Tradeoffs for Hyperbolic Embeddings.
\newblock In \emph{International Conference on Machine Learning (ICML)},
  volume~80 of \emph{Proceedings of Machine Learning Research}, 4457--4466.
  {PMLR}.

\bibitem[{Sarkar(2011)}]{Low_Distor_Delaun_Embed_Trees_Hyper_Plane-Sarkar11}
Sarkar, R. 2011.
\newblock Low Distortion Delaunay Embedding of Trees in Hyperbolic Plane.
\newblock In \emph{International Symposium on Graph Drawing (GD)}, volume 7034
  of \emph{Lecture Notes in Computer Science}, 355--366. Springer.

\bibitem[{Schaefer(2010)}]{Compl_Some_Geomet_Topol_Probl-Schaef10}
Schaefer, M. 2010.
\newblock Complexity of Some Geometric and Topological Problems.
\newblock In \emph{Graph Drawing}, 334--344. Springer Berlin Heidelberg.

\bibitem[{Shimizu, Mukuta, and Harada(2021)}]{Hyper_Neural_Networ-Shimiz21}
Shimizu, R.; Mukuta, Y.; and Harada, T. 2021.
\newblock Hyperbolic Neural Networks++.
\newblock In \emph{International Conference on Learning Representations
  (ICLR)}.

\bibitem[{Suzuki et~al.(2021)Suzuki, Nitanda, wang, Xu, Yamanishi, and
  Cavazza}]{Gener_Bound_Graph_Embed_Using_Negat_Sampl-Suzuk21}
Suzuki, A.; Nitanda, A.; wang, j.; Xu, L.; Yamanishi, K.; and Cavazza, M. 2021.
\newblock Generalization Bounds for Graph Embedding Using Negative Sampling:
  Linear vs Hyperbolic.
\newblock In \emph{Advances in Neural Information Processing Systems
  (NeurIPS)}, volume~34, 1243--1255. Curran Associates, Inc.

\bibitem[{Suzuki, Takahama, and
  Onoda(2019)}]{Hyper_Disk_Embed_Direc_Acycl_Graph-Suzuk19}
Suzuki, R.; Takahama, R.; and Onoda, S. 2019.
\newblock Hyperbolic Disk Embeddings for Directed Acyclic Graphs.
\newblock In \emph{International Conference on Machine Learning (ICML)},
  volume~97 of \emph{Proceedings of Machine Learning Research}, 6066--6075.
  {PMLR}.

\bibitem[{Tifrea, B{\'{e}}cigneul, and Ganea(2019)}]{Poinc_Glove-Tifrea19}
Tifrea, A.; B{\'{e}}cigneul, G.; and Ganea, O. 2019.
\newblock Poincare Glove: Hyperbolic Word Embeddings.
\newblock In \emph{International Conference on Learning Representations
  (ICLR)}. OpenReview.net.

\bibitem[{Tsitsulin et~al.(2018)Tsitsulin, Mottin, Karras, and
  M{\"{u}}ller}]{DBLP:conf/www/TsitsulinMKM18}
Tsitsulin, A.; Mottin, D.; Karras, P.; and M{\"{u}}ller, E. 2018.
\newblock {VERSE:} Versatile Graph Embeddings from Similarity Measures.
\newblock In \emph{International Conference on World Wide Web (WWW)}, 539--548.
  {ACM}.

\bibitem[{van Spengler, Wirth, and Mettes(2023)}]{HypLL-Speng23}
van Spengler, M.; Wirth, P.; and Mettes, P. 2023.
\newblock {HypLL}: The Hyperbolic Learning Library.
\newblock In \emph{International Conference on Multimedia (MM)}, 9676--9679.
  Association for Computing Machinery.

\bibitem[{Voitalov et~al.(2019)Voitalov, van~der Hoorn, van~der Hofstad, and
  Krioukov}]{Scale_Free_Networ_Well_Done-Voital19}
Voitalov, I.; van~der Hoorn, P.; van~der Hofstad, R.; and Krioukov, D. 2019.
\newblock Scale-Free Networks Well Done.
\newblock \emph{Phys. Rev. Res.}, 1: 033034.

\bibitem[{Vyas et~al.(2022)Vyas, Choudhary, Khatir, and
  Reddy}]{GraphZoo-Vyas22}
Vyas, A.; Choudhary, N.; Khatir, M.; and Reddy, C.~K. 2022.
\newblock {GraphZoo}: A Development Toolkit for Graph Neural Networks with
  Hyperbolic Geometries.
\newblock In \emph{The ACM Web Conference (WWW)}. Association for Computing
  Machinery.

\bibitem[{Wang et~al.(2016)Wang, Li, Jin, Xiong, and
  Wu}]{Hyper_Mappin_Compl_Networ_Based_Commun_Infor-Wang16}
Wang, Z.; Li, Q.; Jin, F.; Xiong, W.; and Wu, Y. 2016.
\newblock Hyperbolic Mapping of Complex Networks Based on Community
  Information.
\newblock \emph{Physica A: Statistical Mechanics and its Applications}, 455:
  104--119.

\bibitem[{Yang and Zhou(2024)}]{Hyper_Repres_Deep_Learn-Mengl24}
Yang, M.; and Zhou, M. 2024.
\newblock Hyperbolic Representation and Deep Learning: A Comprehensive
  Collection.
\newblock
  \url{https://github.com/marlin-codes/Awesome-Hyperbolic-Representation-and-Deep-Learning}.

\bibitem[{Zhou et~al.(2023)Zhou, Yang, Xiong, Xiong, and
  King}]{Hyper_Graph_Neural_Networ-Zhou23}
Zhou, M.; Yang, M.; Xiong, B.; Xiong, H.; and King, I. 2023.
\newblock Hyperbolic Graph Neural Networks: A Tutorial on Methods and
  Applications.
\newblock In \emph{International Conference on Knowledge Discovery and Data
  Mining (KDD)}, 5843--5844. Association for Computing Machinery.

\end{thebibliography}

\appendix

\end{document}